\documentclass[manuscript,screen,nonacm]{acmart}
\acmConference[]{} 

\newcommand{\workshopname}{GenAICHI: CHI 2024 Workshop on Generative AI and HCI}
\newcommand\extrafootertext[1]{
    \bgroup
    \renewcommand\thefootnote{\fnsymbol{footnote}}%
    \footnotetext[0]{#1}%
    \egroup
}

\AtBeginDocument{ 
    \fancypagestyle{firstpagestyle}{
        \fancyhf{}
        \fancyfoot[L]{\sffamily\footnotesize \workshopname}%
        \fancyfoot[C]{\sffamily\footnotesize \thepage}
    }
    \fancyhf{}
    \fancyhead[L]{\sffamily\footnotesize\shorttitle}
    \fancyfoot[L]{\sffamily\footnotesize\workshopname}%
    \fancyfoot[C]{\sffamily\footnotesize\thepage}
    \extrafootertext{}
}

\begin{document}

\title{Food Development through Co-creation with AI: bread with a "taste of love"}

\author{Takuya Sera}
\email{takuya-sera@nec.com}
\affiliation{%
  \institution{NEC Corporation}
  \country{Japan}
}

\author{Izumi Kuwata}
\affiliation{%
  \institution{NEC Corporation}
  \country{Japan}
}

\author{Yuki Taya}
\affiliation{%
  \institution{NEC Corporation}
  \country{Japan}
}

\author{Noritaka Shimura}
\affiliation{%
  \institution{NEC Corporation}
  \country{Japan}
}

\author{Yosuke Motohashi}
\affiliation{%
  \institution{NEC Corporation}
  \country{Japan}
}

\begin{abstract}
This study explores a new method in food development by utilizing AI including generative AI, aiming to craft products that delight the senses and resonate with consumers' emotions. The food ingredient recommendation approach used in this study can be considered as a form of multimodal generation in a broad sense, as it takes text as input and outputs food ingredient candidates. This Study focused on producing "Romance Bread," a collection of breads infused with flavors that reflect the nuances of a romantic Japanese television program. We analyzed conversations from TV programs and lyrics from songs featuring fruits and sweets to recommend ingredients that express romantic feelings. Based on these recommendations, the bread developers then considered the flavoring of the bread and developed new bread varieties. The research included a tasting evaluation involving 31 participants and interviews with the product developers.Findings indicate a notable correlation between tastes generated by AI and human preferences. This study validates the concept of using AI in food innovation and highlights the broad potential for developing unique consumer experiences that focus on emotional engagement through AI and human collaboration.
\end{abstract}


\maketitle

\section{INTRODUCTION}
Modern food development is evolving not just by focusing on taste and nutritional value, but also towards stimulating consumers' experiences and emotions. Collaborative products with movies and television programs are an effective strategy aimed at boosting sales by leveraging consumers' emotional connections to specific works. However, traditional food development has relied on human intuition and experience, which is thought to have limitations in creating new experiences. 
In related research, the application of AI in food development, such as the technology of Chef Watson~\cite{varshney2019big}, has played a significant role. Chef Watson was able to create customized granola from more than 50 ingredients, analyzing thousands of ingredient combinations to help consumers explore new flavor combinations and providing data on flavor preferences to contribute to new product development~\cite{walker2024ai}. Moreover, there is research that conducted food development by converting the atmosphere of the era into taste using data from newspaper articles and AI~\cite{sera2020case}.
In this study, we implemented food development using a new approach utilizing AI including generative AI. NEC Corporation~\cite{nec2024} and the long-established bread manufacturer, Kimuraya Sohonten Co.~\cite{kimuraya2024}, jointly created bread, “Romance Bread” ~\cite{necrenai2024}, flavored to resemble the taste of a romantic Japanese television program.

\begin{figure}
    \centering
    \includegraphics[width=1.0\linewidth]{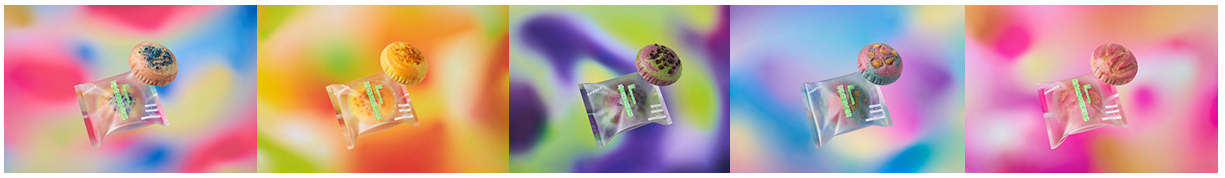}
    \caption{The Romance Bread developed in this case study}
    \label{fig:enter-label}
\end{figure}
We then conducted a tasting survey with participants and interviews with the developers on the developed bread. Based on this feedback, we discuss the potential and future challenges of food development using AI. This study proposes a new method of food development through collaboration between AI and humans, especially exploring the role of AI in creating consumer experiences that emphasize emotions.

\section{DEVELOPMENT BACKGROUND}
Recent market research targeting Japanese youth, who are the primary audience for this project, revealed a trend of distancing from romantic relationships. Therefore, we decided to conduct food development themed around love by choosing a romantic Japanese television program as our subject.
Kimuraya Sohonten Co., with a history of over 150 years, has traditionally sold its products through department stores and supermarkets. As a result, acquiring new customers and increasing brand awareness among younger consumers have been challenges.
NEC has previously proposed new approaches to food development using AI technology, including chocolate, craft beer, and pudding.

\section{TECHNOLOGIES}
NEC Enhanced Speech Analysis (NESA)~\cite{necspeech2024} combines voice recognition and analysis technologies, and in this study, it was used for transcribing conversations within the program. 
Data Enrichment (DE)~\cite{necenrichment2024} transforms text (e.g., audience reactions or conversations within the program) into a more analyzable form.~\cite{yano2021zero} This technology allows for the interpretation of text meaning and scoring of keywords expressing emotion.
NEC cotomi~\cite{neccotomi2024} is a lightweight Large Language Model (LLM) developed by NEC, featuring high Japanese language performance. By implementing unique innovations, this LLM achieves high performance while keeping the number of parameters to 13 billion.

\section{ROMANCE BREAD DEVELOPMENT PROCESS}
By calculating emotion vectors from romance scenes extracted from a romantic Japanese television program and emotion vectors from ingredients extracted from song lyrics, we recommended a list of ingredients with high similarity in emotion vectors for each romance scene. The romance scenes are categorized into five themes: "First Encounter," "Date," "Jealousy," "Broken Heart," and "Mutual Love." Based on these ingredient lists, the bread developers considered combinations and appearances of ingredients to develop five types of bread.

\subsection{Ingredients Recommendation}
Process1 Conversation Analysis: Using NESA, we extracted 15 hours of conversation data from participants in a romantic Japanese television program as text and categorized them into five scenes. The number of records for each scene was as follows: "First Encounter" had 1,073 records, "Date" had 2,855 records, "Jealousy" had 314 records, "Mutual Love" had 140 records, and "Broken Heart" had 206 records.
By providing each conversation sentence with 32 types of emotion tags ~\cite{plutchik2001nature}, we calculated a 32-dimensional emotion vector for each sentence. Then, we aggregated emotion vectors for each scene. As a result, we were able to create emotion vectors for the five scenes.
Process2 Lyric Analysis: Similarly to the romance scenes, we wanted to assign emotion vectors to ingredients and decided to use song lyrics as the text data to express the emotions of ingredients, since ingredients often appear in lyrics and are used to convey emotions. Therefore, we extracted about 35,000 songs from a database of approximately 1 million Japanese song lyrics, which include 183 types of food items like fruits and sweets. Like with the romance scenes, we calculated a 32-dimensional emotion vector for each lyric and aggregated emotion vectors for each ingredient. As a result, we created emotion vectors for 183 ingredients.
Process3 Ingredients Recommendation: By obtaining the cosine similarity between the emotion vectors of the romance scenes and the emotion vectors of the ingredients, we listed the top 50 ingredients that closely match the emotional distribution for each romance scene, expressing the emotions of love, and recommended them to the bread developers.

\subsection{Bread Development}
Bread developers at Kimuraya Sohonten Co., who have developed numerous steamed breads in the past, selected compatible combinations of ingredients from the 50 kinds guided by AI to develop five types of "Romance Bread," recreating love as a taste. For example, the "First Encounter" Bread was made with the following ingredients: Dough: Cotton Candy, Marble: Apple, Topping: Blue Crunch. The "Date" Bread was created using Dough: Lime, Marble: Persimmon, Topping: Orange Peel.

\subsection{Product Description Generation}
Based on the romance scenes and ingredients used, we considered prompts for generating product descriptions and conducted text generation using NEC cotomi. Based on the results, manual corrections were made to create product descriptions used on packaging and special websites. For instance, regarding "First Encounter," the results from NEC cotomi were manually revised, resulting in the Product Description: "The sweet scent of 'apple' ticked the heart on a breeze of 'cotton candy.' Is this a coincidence? The 'crunch' makes one feel the playfulness of fate, hinting at a love about to begin."

\section{EVALUATION}
\subsection{Tasting Evaluation}
The tasting evaluation was conducted with 31 participants aged between 20 and 50. They tasted the five types of bread without being told the titles and were asked to identify which title each bread represented. 

\begin{table}[h]
\centering
\caption{Confusion Matrix of Evaluation Results}
\label{tab:confusion_matrix}
\begin{tabular}{|c|c|c|c|c|c|}
\hline
{Actual} & \multicolumn{5}{c|}{Evaluator's Answer} \\ \cline{2-6} 
 & First Encounter & Date & Jealousy & Mutual Love & Broken Heart \\ \hline
First Encounter & 15 & 5 & 4 & 2 & 5 \\ \hline
Date & 3 & 11 & 3 & 5 & 9 \\ \hline
Jealousy & 5 & 0 & 21 & 1 & 4 \\ \hline
Mutual Love & 5 & 7 & 1 & 10 & 8 \\ \hline
Broken Heart & 6 & 5 & 3 & 13 & 4 \\ \hline
\end{tabular}
\end{table}

The accuracy rate was 43.8\%.This result significantly exceeds the expected accuracy of 20\% if selections were made randomly, suggesting that there is a certain similarity between taste indicators generated by AI and human sensibilities. Upon observing the results for each flavor, the highest precision was for "Jealousy" flavor at 65.6\%. This indicates that it was possible to accurately capture specific emotions with concrete taste experiences. Conversely, the lowest precision was for "Mutual Love" flavor at 26.7\%. Many participants mistakenly answered it as "Date," suggesting that the emotions associated with "Mutual Love" and "Date" are close, making it challenging to distinguish and express them with ingredients.
Furthermore, participants provided positive feedback such as:
"I would love to buy a variety of these breads as a couple or family, share them, and enjoy them together."
"The bread itself was quite delicious. I thought it would be interesting to extend the same taste and names to other foods and beverages."
These comments suggest that not only the taste of the bread but also its concept could propose new enjoyable experiences to consumers.

\subsection{Interviews with Developers}
The developers considered combinations and hues of flavors for the bread based on the ingredient list recommended by AI. They shared comments post-development such as:
\begin{itemize}
    \item "We particularly focused on the color, pursuing a design where the image of emotions and the taste of the ingredients are connected."
    \item "In developing the 'Jealousy' flavor bread, it was especially challenging to adjust the colors to express the emotions."
    \item "Utilizing AI allowed us to challenge ourselves with untried ingredients and combinations, discovering new possibilities in bread-making."
\end{itemize}

\section{DISCUSSION}
This study's findings indicate a synergy between AI-generated taste indicators and human preferences, with notable precision in the "Jealousy" flavored bread at 65.6\%, showcasing AI's potential in creating novel taste experiences that appeal to consumer emotions. Developers' insights reveal that AI collaboration fosters innovative flavor combinations and challenges, like color adjustments to express emotions, thus pushing the boundaries of creativity in food development.
The AI used in this study was limited to ingredient recommendations and generating product descriptions. However, combining various generative AI technologies, including image generation AI, could lead to innovative ideas in food appearance and packaging design. Additionally, this study proposed a one-way flow where developers create products based on AI recommendations. A more refined product development could be expected through interactive exchanges between developers and AI. For example, integrating a bidirectional communication where developers input feedback into AI after creating prototypes, and AI uses this to propose further optimized ingredients or designs. This process is expected to blend AI suggestions with human intuition and experience, creating products that better meet consumer expectations.

\section{CONCLUSION}
Through this study, we confirmed that utilizing AI in food development offers new value. Proposals by AI for ingredients highly align with human sensibilities, successfully providing new taste experiences that consumers seek. The future of food development is anticipated to grow significantly by incorporating a wider range of generative AI technologies. This co-creation process might potentially accelerate innovation in the food industry and could contribute to delivering new value to consumers. Collaboration between AI and humans could play an important role in shaping the future of food development, possibly leading to the creation of new consumer experiences.

\newpage
\bibliographystyle{unsrt}
\bibliographystyle{ACM-Reference-Format}
\bibliography{example-references}

\begin{thebibliography}{10}

\bibitem{varshney2019big}
L.~R. Varshney, F.~Pinel, K.~R. Varshney, D.~Bhattacharjya, A.~Schörgendorfer, and Y.~.~M. Chee.
\newblock A big data approach to computational creativity: The curious case of chef watson.
\newblock {\em IBM Journal of Research and Development}, 63(1):7:1--7:18, 2019.

\bibitem{walker2024ai}
J.~Walker.
\newblock Ai in food processing use cases and applications that matter.
\newblock \url{https://emerj.com/ai-sector-overviews/ai-in-food-processing/}, 2024.
\newblock Accessed February 5, 2024.

\bibitem{sera2020case}
Takuya Sera, Sayaka Izukura, Izumi Hashimoto, Takashi Motegi, and Yosuke Motohashi.
\newblock A case study of food production using artificial intelligence.
\newblock In {\em Proceedings of the 2020 Symposium on Emerging Research from Asia and on Asian Contexts and Cultures}, AsianCHI '20, 2020.

\bibitem{nec2024}
{NEC Corporation}.
\newblock {NEC Corporation}.
\newblock \url{https://www.nec.com/}, 2024.
\newblock Accessed February 5, 2024.

\bibitem{kimuraya2024}
{Kimuraya Sohonten Co.}
\newblock {Kimuraya Sohonten Co.}
\newblock \url{https://www.kimuraya-sohonten.co.jp/}, 2024.
\newblock In Japanese. Accessed February 5, 2024.

\bibitem{necrenai2024}
{NEC Corporation}.
\newblock {Ren AI Pan}.
\newblock \url{https://www.nec.com/en/press/202401/global_20240117_01.html}, 2024.
\newblock Accessed February 5, 2024.

\bibitem{necspeech2024}
{NEC Corporation}.
\newblock {NEC Enhanced Speech Analysis}.
\newblock \url{https://jpn.nec.com/press/202203/20220307_01.html}, 2024.
\newblock In Japanese. Accessed February 5, 2024.

\bibitem{necenrichment2024}
{NEC Corporation}.
\newblock {Data Enrichment}.
\newblock \url{https://jpn.nec.com/solution/dataenrichment/index.html}, 2024.
\newblock In Japanese. Accessed February 5, 2024.

\bibitem{yano2021zero}
Taro Yano, Kunihiro Takeoka, and Masafumi Oyamada.
\newblock Zero-shot text classification using hierarchical novelty detection.
\newblock In {\em The 35th Annual Conference of the Japanese Society for Artificial Intelligence}, 2021.

\bibitem{neccotomi2024}
{NEC Corporation}.
\newblock {cotomi}.
\newblock \url{https://jpn.nec.com/press/202312/20231215_02.html}, 2024.
\newblock In Japanese. Accessed February 5, 2024.

\bibitem{plutchik2001nature}
R.~Plutchik.
\newblock The nature of emotions: Human emotions have deep evolutionary roots.
\newblock {\em American Scientist}, 89(4):344--350, 2001.

\end{thebibliography}

\end{document}